%% file: root.tex
\documentclass[letterpaper, 10 pt, conference]{ieeeconf}  
\usepackage{hyperref}
\usepackage{svg}

\IEEEoverridecommandlockouts                              

\usepackage{caption}
\usepackage[utf8]{inputenc} 

\usepackage[T1]{fontenc}    
\usepackage{hyperref}       
\usepackage{url}            
\usepackage{booktabs}       
\usepackage{amsfonts}       
\usepackage{nicefrac}       
\usepackage{microtype}      
\usepackage{xcolor}         
\usepackage{multirow}
\usepackage{graphicx}
\usepackage{makecell}

\usepackage[T1]{fontenc}
\usepackage{lmodern}  
\usepackage{graphicx}
\usepackage{amssymb}
\usepackage{booktabs}
\usepackage{multirow}
\usepackage{multicol}
\usepackage{caption} 
\usepackage{longtable} 
\usepackage{color}
\usepackage{pifont}%
\usepackage{array}
\usepackage{colortbl}

\usepackage{amsmath,amssymb,amsfonts} 
\usepackage{mathtools}                 
\usepackage{bm}                        
\usepackage{siunitx}                   
\usepackage{microtype}                 

\usepackage{newtxtext}
\usepackage[dvipsnames]{xcolor}

\title{\LARGE \bf
\textcolor{RoyalBlue}{OWMDrive:} Causality-Aware End-to-End Autonomous Driving via 4D \textcolor{RoyalBlue}{O}ccupancy \textcolor{RoyalBlue}{W}orld \textcolor{RoyalBlue}{M}odel
}
\author{
    Junjie Cheng\textsuperscript{1,2,*}, Ruiqi Song\textsuperscript{2,3,4,*}, Ye Wu\textsuperscript{1,2}, Nanxing Zeng\textsuperscript{1,2}, Ximiao Li\textsuperscript{1,2}, Yunfeng Ai\textsuperscript{1,2,\textdagger}\\
    \thanks{
        This work was supported by the Key Research and Development Program of Shaanxi Province (2024CY2-GJHX-49) and the Key Research and Development Program of Xinjiang Uyghur Autonomous Region (Grant No.\ 2025B01001).
    }
       \thanks{* These authors contributed equally to this work.}
    \thanks{\textdagger\ \textbf{Corresponding author}:Y. Ai. \texttt{aiyunfeng@ucas.ac.cn}}
    \thanks{
        \textsuperscript{1} The School of Artificial Intelligence, University of Chinese Academy of Sciences, Beijing 100049, China \texttt{\{chengjunjie25, wuye23, zengnanxin24, liximiao26\}@mails.ucas.ac.cn}
    }
    \thanks{\textsuperscript{2} Waytous Inc., Qingdao 266109, China}
    \thanks{
        \textsuperscript{3} The State Key Laboratory of Multimodal Artificial Intelligence Systems, Institute of Automation, Chinese Academy of Sciences, Beijing 100190, China \texttt{ruiqi.song@ia.ac.cn}
    }
    \thanks{
    \textsuperscript{4} The College of Surveying and Geo-Informatics, Tongji University, Shanghai 200092, China.
    }
}

\begin{document}
\maketitle
\thispagestyle{empty}
\pagestyle{empty}

\begin{abstract}
Autonomous driving systems are steadily moving toward end-to-end paradigms to mitigate the limited adaptability of rule-based pipelines in complex traffic environments. However, most existing learning-based methods still make decisions from static representations of the current scene, without explicit future rollouts or modeling of the temporal causal dynamics in traffic interactions. This limitation often results in unstable or overly conservative planning under high-uncertainty conditions, such as occlusions and unexpected events. To overcome these challenges, we introduce OWMDrive, a generative end-to-end driving framework built upon an Occupancy World Model for multi-step 3D occupancy forecasting, which serves as a conditional prior to guide diffusion-based planning. Conditioned on both current observations and predicted future states, the planner iteratively refines trajectory candidates to generate a reinforced driving trajectory. By explicitly modeling scene evolution over future horizons, OWMDrive captures key spatiotemporal causal dependencies, which leads to more foresighted and robust trajectory generation. Extensive experiments demonstrate that OWMDrive significantly improves planning reliability and safety, especially in challenging and partially observable driving scenarios.

\end{abstract}

\section{INTRODUCTION}
The field of autonomous driving has attracted 
considerable attention\cite{yu2025end,sun2025sparsedrive}, 
driven by rapid advances in perception modules and trajectory planning. A growing number of data-driven methods learn driving behaviors directly from raw sensor inputs, such as RGB-D images\cite{yang2020rgb} and point clouds\cite{8954311}. In contrast, conventional rule-based systems\cite{bouchard2022rule} often struggle to handle complex and dynamic real-world environments, limiting their robustness and scalability across diverse scenarios. By leveraging large-scale datasets collected under diverse driving conditions, data-driven methods\cite{amini2020learningdatadriving1} can learn richer behavioral representations.

Most contemporary data-driven methods rely primarily on perceptual cues derived from current and past observations\cite{TransfuserKashyapChitta,liao2025diffusiondrive}, such as object detections and the relative spatial relationships among surrounding agents. 
However, such reasoning is inherently limited in capturing the underlying causal structure of scene dynamics, often leading to suboptimal or unsafe planning under challenging conditions such as severe occlusions. Furthermore, despite 
the availability of extensive real-world driving datasets, performance 
is still hindered by data quality and diversity. Underrepresented or complex real-world scenarios\cite{huang2024versatile} often suffer from limited generalization, resulting in inadequate path planning.

\begin{figure}[t]
\centering
\includegraphics[width=\columnwidth]{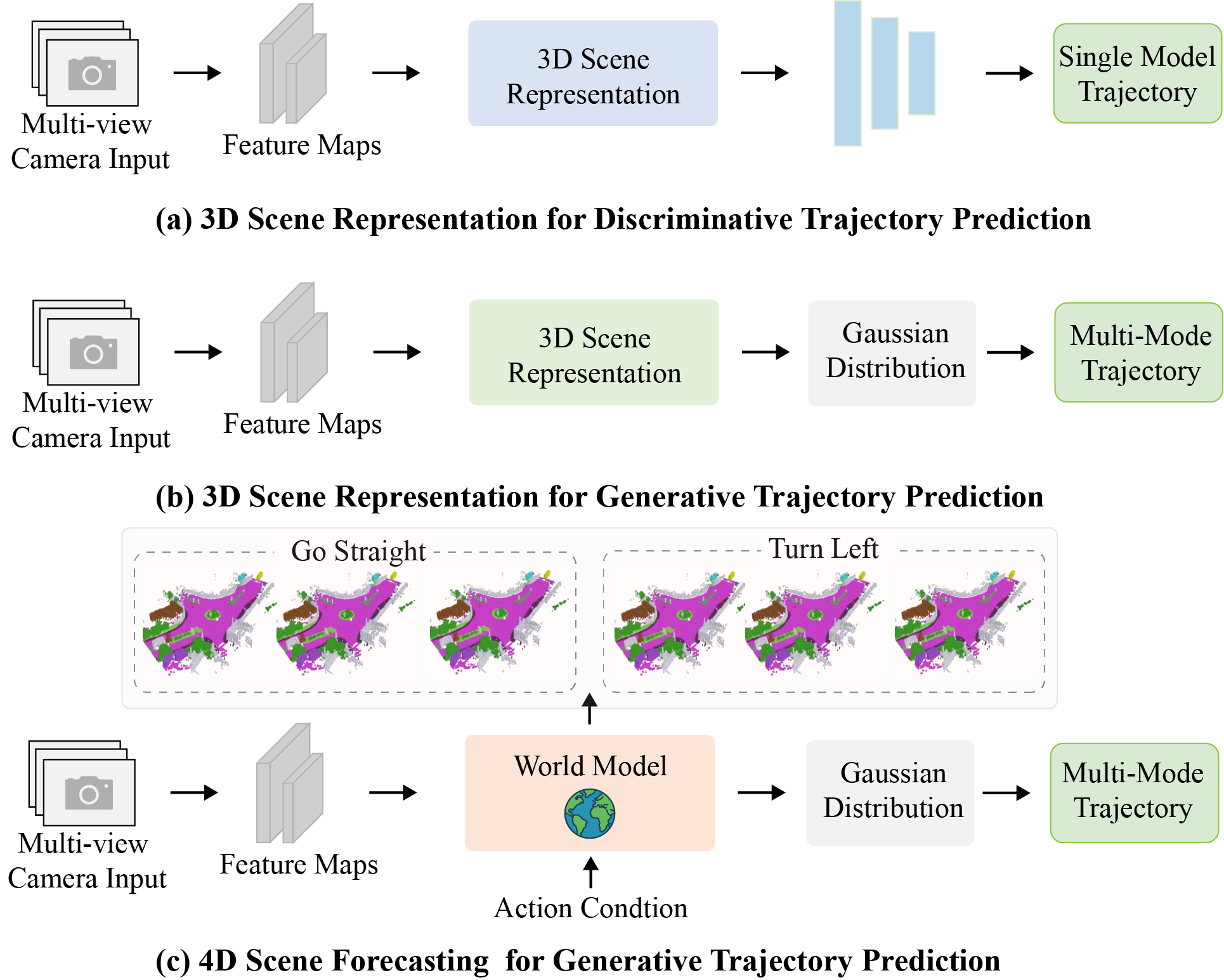}
\caption{Evolution of generative end-to-end autonomous driving paradigms.}
\vspace{-20pt}
\label{fig:example}
\end{figure}

Our solution to these challenges is a diffusion-based end-to-end driving framework that incorporates a predictive world model to explicitly model future scene evolution. 
The proposed architecture consists of two key components: (1) an occupancy-based world model that forecasts future 3D scene occupancy from current and historical observations, and (2) a planning module that integrates perception with a diffusion-based trajectory generator. The planner conditions on multi-step future predictions provided by the world model, and iteratively denoises trajectory samples to produce the final plan. 
To improve trajectory quality and stability, the diffusion process is further optimized via reinforcement learning, encouraging safe and goal-consistent exploration. 
By providing explicit foresight into scene dynamics, the world model enables causality-aware conditioning, allowing the diffusion model to generate trajectories that better handle complex and partially observable environments.

Our key contributions are threefold:
\begin{itemize}
    \item We propose OWMDrive, a reinforced generative end-to-end driving framework that explicitly conditions diffusion-based planning on future scene rollouts, supporting causality-aware trajectory planning.
    \item We develop an occupancy world model to forecast multi-step 3D occupancy evolution and serve as a conditional prior for the planner, capturing key spatiotemporal causal dynamics of traffic interactions.
    \item Systematic open-loop and closed-loop evaluations are conducted on standard benchmarks, demonstrating improved planning reliability and safety, particularly under challenging and partially observable scenarios.
\end{itemize}

\section{Related Work}
\subsection{World Models for Autonomous Driving.}
World models learn environmental dynamics to predict future states and have become a cornerstone for enhancing the reasoning capabilities of autonomous driving systems. Early applications often focused on latent-space dynamics modeling in simplified environments\cite{zhang2024bevworldlatentspace1,popov2024mitigatinglatentspace2}. However, recent research has shifted toward learning explicit, high-fidelity representations of the 3D world, primarily through 3D occupancy grids\cite{gao2024vistaocc1,gu2024domeocc2,xu2025occ,zheng2024occworld}. This representation captures the geometry and semantics of the scene, which provides a rich, unified format for both static and dynamic elements.

A significant line of research focuses on predicting future 3D scene occupancy from historical sensor observations. For example, OccWorld\cite{zheng2024occworld} introduced a 3D occupancy world model that learns spatiotemporal dynamics in bird's eye view (BEV) features, enabling future environment forecasting. 
However, its predictions may suffer from a lack of temporal consistency. To address this issue, OccSora\cite{wang2024occsora} extended this framework into full 4D forecasting, formulating the problem as generative simulation. 
To tackle the efficiency challenge, OccProphet \cite{chen2025occprophet} proposes an efficient observer-forecaster-refiner framework to improve the accuracy and efficiency of camera-only 4D occupancy forecasting. Several works further explore integrating predictive world models into the planning pipeline. 
For example, DriveWorld\cite{yang2025driving} develops a perception-centric system that directly leverages 4D occupancy forecasts for downstream planning. In addition, $I^2$-World\cite{liao20252} proposes an Intra-Inter tokenization strategy designed to balance the compactness inherent in 3D tokenization against the expressive capacity of 4D forecasting. 
Occ-LLM\cite{xu2025occ} incorporates a large language model (LLM) to process occupancy-based world model outputs, enabling language-grounded reasoning for complex driving scenarios. 
These works collectively indicate a shift from static 3D perception toward dynamic 4D world modeling to support predictive decision-making. 
\subsection{Generative End-to-End Autonomous Driving}
The core objective of end-to-end planning is to derive driving trajectories from sensor observations.
Recently generative models\cite{zheng2024genad,insightdrive} have gained increasing attention for this task. Unlike deterministic approaches that predict a single optimal path, generative methods model a distribution over plausible future trajectories, naturally capturing the inherent multi-modality of driving behavior.

Among generative approaches, LLMs and diffusion models have shown 
strong potential, owing to their proficiency in capturing intricate data distributions and synthesizing a wide range of plausible outputs.

For example, DriveGPT4\cite{xu2024drivegpt4} leverages a multimodal language model to process video inputs 
formulating driving as a conditional sequence generation problem for control signal prediction. 
Concurrently, GPVL\cite{li2025generative} introduces an end-to-end autonomous driving framework that integrates 3D visual representation learning with language-guided generative planning for trajectory prediction. 
Nevertheless, methods based on LLMs 
are computationally intensive, which severely hinders their feasibility for real-time onboard execution and potentially introduces reasoning inconsistencies in complex scenarios. 
\begin{figure*}[t]
    \centering
    \includegraphics[width=1\textwidth]{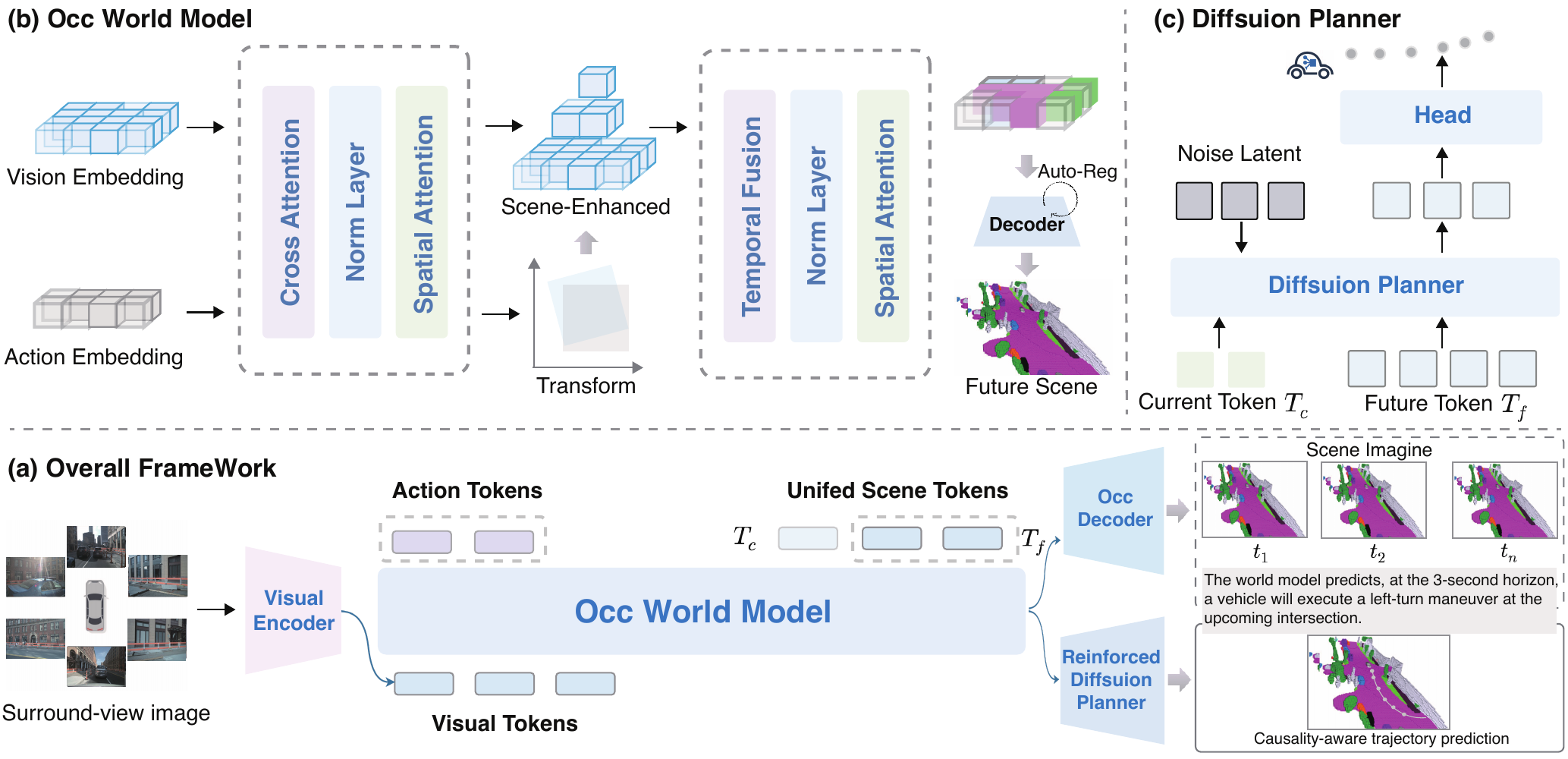}
    \caption{\textbf{Overview of OWMDrive framework.} 
(a) Surround-view images are encoded into visual tokens and processed by an Occupancy World Model to produce current and future scene tokens $(T_c, T_f)$. 
(b) The world model forecasts multi-step future 3D occupancy via spatiotemporal fusion and autoregressive decoding. 
(c) Conditioned on $(T_c, T_f)$, a diffusion planner iteratively refines trajectory latents to predict waypoints.}
    \label{fig:full_algorithm}
    \vspace{-15pt}
\end{figure*}
In contrast to LLM-based approaches, diffusion models iteratively refine a noisy trajectory into a safe and feasible path. DiffusionDrive\cite{liao2025diffusiondrive} integrates perception and planning into a unified architecture, where a diffusion model generates diverse and context-aware trajectories directly from sensor inputs. Similarly, Diffusion-AD\cite{wang2025diffad} 
recasts end-to-end driving into a conditional generation task within the BEV space. It employs a conditional diffusion model 
to construct a unified BEV representation that jointly encapsulates environmental perception, state prediction, and motion planning.

Existing diffusion-based approaches condition trajectory generation on perceptual features of the current scene. 
In contrast, our work conditions trajectory generation not only on the current scene representation but also on multi-step future scene predictions from a 4D occupancy world model, incorporating predicted environmental dynamics into the planning process.
\section{Method}
\subsection{Preliminary} 
In conditional diffusion models, the conditioning signal directly guides the reverse denoising process. 
Incorporating predicted future states into the conditioning information can further stabilize the training process and lead to more robust planning outcomes. 
Specifically, the planning process is conditioned 
on a sequence of predicted future environmental states at time $t$, 
denoted as $c_t=s_{t+1:t+f}$.


We formulate trajectory planning in end-to-end autonomous driving as a conditional generation problem, where a future trajectory is generated conditioned on scene information. Conditional diffusion models learn to model the trajectory distribution through a forward diffusion process that progressively corrupts data and a conditional reverse process that denoises samples. The forward process is defined as:
\begin{equation}
    q(\tau_{1:K}|\tau_0)=\prod_{k=1}^Kq(\tau_k|\tau_{k-1})
\end{equation}
where $\tau_k$ is the data sampled with noise at diffusion timestep $k$. During the denoising process, incorporating relevant information from the original data distribution as a condition can effectively guide the model to approximate the distribution. 
It can be formulated as:
\begin{equation}
    q(\tau_{k-1}|\tau_k,\tau_0, c_t)=\frac{q(\tau_{k}|\tau_{k-1},c_t)q(\tau_{k-1}|\tau_0,c_t)}{q(\tau_k|\tau_0,c_t)}
\end{equation}



\subsection{4D Scene Forecasting with World Models} 
\noindent\textbf{Semantic Scene Tokenizer.}
In end-to-end autonomous driving, comprehensive scene representation is fundamental to downstream planning and decision-making. 
To effectively capture dynamic scene evolution, we encode high-dimensional 3D occupancy representations into a compact discrete latent space $\mathcal{Z}$. 
This transformation is achieved via a representation learning framework that integrates spatial compression and temporal alignment. Specifically, the spatial compression component performs a multi-scale residual quantization that hierarchically encodes spatial structures at each physical time step $t$, which enables the preservation of fine-grained geometric details.
The temporal component explicitly models dynamic changes over time by maintaining a historical queue, which aligns features across different times. 

The spatial compression module takes the 3D occupancy representation $O_t$ as input and encodes it into a latent representation using a multi-scale residual quantization framework built upon VQ-VAE\cite{lee2022autoregressiveVQVAE} and VAR\cite{tian2024visualVAR}. A ResNet-based decoder reconstructs $O_t$ from the quantized latent codes.  
Specifically, the encoder first maps $O_t$ to a feature map $F_t$ and progressively downsamples to produce $Z$ multi-scale feature maps $F_t^1,\dots,F_t^{Z}$, where $F_t^z\in \mathbb{R}^{H_z \times W_z \times C}$. $H_z$ and $W_z$ denote the spatial dimensions, and $C$ denotes the channel dimension. 
A shared learnable codebook $\mathcal{C}$ containing $N$ code vectors of dimension $C$ is employed for quantization.
\begin{equation}
        Q_t^z(i) = \operatorname*{argmin}_{c\in \mathcal{C}} || R_t^z(i)-c||_2
\end{equation}
At each scale, the quantizer encodes the residual between the current feature map and the upsampled reconstruction from the previous scale. The residual is initialized as $F^1_t$. This coarse-to-fine residual quantization progressively refines the representation across scales. The process is formulated as:
\begin{equation}
        R_t^{z+1}=F_t^{z+1}-Up(Q_t^z)
\end{equation}
\begin{equation}
        \hat{F}_t=\sum_{z=1}^Z(Q_t^z)
\end{equation}
where $||\cdot||$ is the L2 norm, $Up$ denotes an upsampling operation and $\hat{F}_t$ represents the aggregated features from ${F}_t$. 
This hierarchical quantization mitigates detail loss commonly encountered in single-scale approaches, thereby enhancing the accuracy and stability of the reconstructed 3D occupancy scenes.

The temporal module explicitly models temporal evolution. To ensure alignment across time, we maintain a memory queue $M_t=\{F_{t-H},\dots,F_{t-1}\}$ containing $H$ historical feature maps. For each historical timestep $t-h$, a transformation matrix ${T}_{t-h}^t$ is applied to align the features with the current coordinate system, denoted as $F_{t-h}'$. 
The aligned historical features are quantized similarly to the spatial part. The beginning residual $R_t^{(0)}$ is set as $\hat{F}_t$ 
and a temporal token map is generated through an iterative process over $H$ steps: 
\begin{equation}
        Q_t^{(h)}(i)=\operatorname*{argmin}_{c\in \mathcal{C}} || (R_t^{(h-1)}+F_{t-h}')(i)-c||_2
\end{equation}
\begin{equation}
        R_t^{(h)}=R_t^{(h-1)}-Q_t^{(h)}
\end{equation}
The final output $\hat F_t$ integrates 
both spatial structure and dynamic information: 
\begin{equation}
    \hat F_t=\hat F_t + \sum_{h=1}^HQ_t^{(h)}
\end{equation}
Through residual aggregation, motion patterns can be effectively captured with only a few convolutional layers, which significantly reduces the computational cost associated with 4D modeling.

\noindent\textbf{Auto-Regressive Spatial-Temporal  Modeling.}
To effectively exploit both present and anticipated future scene information during the diffusion denoising process, the stability and fidelity of foresighted scene representations become crucial.
Therefore, we introduce an auto-regressive world model that operates on tokenized scene representations to forecast future scene token sequences. 
Unlike purely decoder-based architectures, we adopt an encoder-decoder paradigm to effectively decouple spatial context aggregation from temporal dynamics modeling. 
Given the current scene representation $\hat F_{t}$ and historical context $\hat F_{t-h:t-1}$, the model iteratively predicts future scene representations $\hat F_{t+1:t+f}$. 
The encoder hierarchically aggregates spatial features from the current feature map while predicting a transformation matrix for temporal guidance. We use Spatial Self-Attention (SSA) with deformable mechanisms to aggregate spatial context. For each query feature $q_{x,y}$ in $\hat F_t$, the attention weights are computed adaptively by sampling a sparse set of key-value pairs from the token map. The output is formulated as:
\begin{equation}
\psi(q_{x,y},v)=\sum_{m=1}^M\omega_mv(p+\Delta p_m),
\end{equation}
where $\psi$ represents the cross-attention operation, $M$ denotes the number of sampling locations and $v$ denotes the value derived from $\hat F_t$. The spatial coordinate $p=(x,y)$ represents the query location, while $\Delta p_m$ are learned offsets that adaptively focus on semantically informative regions (e.g., moving objects). The attention weights $\omega_m$ are computed from the similarity between the query feature and the sampled key features. This deformable attention reduces computational overhead compared to global attention. To capture hierarchical spatial representations, the encoder applies progressive downsampling to produce multi-scale feature maps $\bar{F}_{t}^1,\dots,\bar{F}_{t}^l$, which are integrated through a Feature Pyramid Network (FPN). Concurrently, historical ego-motion information $P_t^0$ interacts with spatial tokens through cross-attention, yielding a refined planning feature $P_t^l$. This feature is projected via a multilayer perceptron (MLP) to regress the transformation matrix $T_{t+j}^{t+j+1}\in \mathbb{R}^{4\times 4}$, which maps scene representation from timestep $t+j$ to $t+j+1$ and serves as a compact representation of scene dynamics.

The decoder generates future tokens in an auto-regressive manner by conditioning on the predicted transformation matrix and historical context. Specifically, the transformation matrix is first embedded into the latent space of scene representations and combined with the spatial features $\bar{F}_{t+j}$ to form a conditioned representation. This representation is then processed with spatial self-attention to maintain frame-level spatial consistency. Meanwhile, temporal dependencies are incorporated through a lightweight temporal fusion module that aggregates historical tokens $F_{t-h:t-1}$. Concretely, historical features are concatenated along the channel dimension and passed through a single MLP layer, enabling efficient temporal modeling without introducing excessive parameters. The resulting features are finally integrated to produce the predicted future scene representation $\hat F_{t+j+1}$.

\subsection{Reinforced Causality-Aware Diffusion}

To effectively leverage both current and predicted future scene information, we design the trajectory planner as a conditional diffusion model that generates safe and efficient trajectories by jointly reasoning over present and anticipated future states.
Conventional diffusion-based planners often start denoising from pure Gaussian noise, requiring many iterations and suffering from mode collapse in dynamic scenes.
To alleviate these issues, we introduce a causality-aware diffusion planner conditioned on rich spatiotemporal features derived from 
multi-step future predictions provided by the world model.

\noindent\textbf{4D Scene Conditioning.}
Let $\bar{s}_t^k$ denote the latent trajectory representation at denoising step $k$, and $\hat F_{t+1:t+f}$ be the $f$-step predicted scene sequence from the world model.
We enhance the diffusion process by embedding foresighted and causal information via a cross-attention fusion:
\begin{equation}
\bar{s}_t^k \leftarrow \mathrm{LN}\big(\bar{s}_t^k + \psi(\bar{s}_t^k, \hat F_{t+1:t+f})\big),
\end{equation}
where $\psi(\cdot)$ denotes a temporal cross-attention operator and $\mathrm{LN}$ is layer normalization.
This operation explicitly couples current features with predicted future dynamics, producing causality-aware latent states that align temporal evolution across predicted scenes.

During training, the model learns a mapping
\begin{equation}
p_\theta(\tau \mid \bar{s}_t, \hat F_{t+1:t+f}) = \mathcal{D}_\theta(\tau_T),
\end{equation}
where $\tau_T$ is the noised trajectory and $\mathcal{D}_\theta$ is the learned denoising function, guiding generation toward the optimal action sequence given both present and future context:
\begin{equation}
\tau^* = \arg\max_\theta \; p_\theta(\tau \mid \bar{s}_t, \hat F_{t+1:t+f}).
\end{equation}

\input{tables/open}

\noindent\textbf{Reinforced Diffusion for Trajectory Planning.}
Instead of sampling from a standard Gaussian prior, we define a fixed Gaussian mixture over a small set of multi-modal reference trajectories $\mathcal{A} = \{a_1, \dots, a_K\}$ obtained by clustering training samples.
At the $k$-th step, trajectory noise is scheduled as:
\begin{equation}
\tau_k = \sqrt{\alpha_k}\, a_i + \sqrt{1-\alpha_k}\,\epsilon, \quad \epsilon \sim \mathcal{N}(0, I),
\end{equation}
where $\alpha_k$ controls the noise level and $a_i\in \mathcal{A}$ is sampled by mode probability $p_i$. During reinforcement learning (RL)-enabled training we further apply scale-adaptive exploration $\tau'=(1+\epsilon_{mul})\tau$ to keep exploratory trajectories smooth and coherent across short and long horizons. The denoising objective is augmented with a trajectory-level RL term, and the standard denoising loss is:
\begin{equation}
\mathcal{L}_{\text{diff}} =
\mathbb{E}_{(\tau, \hat F),\, \epsilon,\, k}
\!\left[
\left\|
\epsilon - 
\epsilon_\theta(\tau_k, \hat F_{t+1:t+f}, k)
\right\|_2^2
\right],
\end{equation}
where $\epsilon_\theta$ predicts the added noise.
The loss is combined with GRPO-based policy gradients, which compute group-relative advantages within each reference and stabilize learning by clipping negative advantages and heavily penalizing collisions. 
A coarse-to-fine mode selector 
then picks the most goal-aligned trajectory. We initialize from $\mathcal{A}$, preserving semantic priors while incorporating exploration and GRPO. This allows the planner to rapidly converge to high-quality, diverse trajectories within very few steps, while simultaneously enhancing safety and mode consistency.

\noindent\textbf{GRPO Reward Design.}
To optimize the diffusion planner with trajectory-level feedback, we define the reward
using closed-loop planning metrics rather than a differentiable regression loss. For each
denoised trajectory $\tau_i$, we evaluate five normalized driving-quality signals,
including no at-fault collision $r_i^{\mathrm{NC}}$, drivable-area compliance
$r_i^{\mathrm{DAC}}$, time-to-collision safety $r_i^{\mathrm{TTC}}$, ego progress
$r_i^{\mathrm{EP}}$, and comfort $r_i^{\mathrm{C}}$. The final scalar reward is computed
as
\begin{equation}
R(\tau_i) =
r_i^{\mathrm{NC}} r_i^{\mathrm{DAC}}
\cdot
\frac{
5 r_i^{\mathrm{TTC}} +
5 r_i^{\mathrm{EP}} +
2 r_i^{\mathrm{C}}
}{12}.
\end{equation}
Here, $r_i^{\mathrm{NC}}$ and $r_i^{\mathrm{DAC}}$ act as multiplicative safety gates,
such that trajectories causing at-fault collisions or leaving the drivable area receive
strongly suppressed rewards. The remaining terms measure risk awareness, route progress,
and motion smoothness, with larger weights assigned to safety-critical time-to-collision
and progress.

For GRPO optimization, we generate a group of trajectory samples under each reference mode
and compute their relative advantages according to the above reward. Given the rewards $
\{R(\tau_i)\}_{i=1}^{G}$ within a group, the normalized advantage is
\begin{equation}
\hat{A}_i =
\frac{
R(\tau_i) - \mu_R
}{
\sigma_R + \epsilon_c
},
\quad
\mu_R = \frac{1}{G}\sum_{i=1}^{G} R(\tau_i),
\end{equation}
where $\sigma_R$ is the standard deviation of group rewards and $\epsilon_c$ is a small
constant for numerical stability. To avoid encouraging unsafe exploratory samples, we
further apply safety-aware advantage truncation:
\begin{equation}
A_i =
\begin{cases}
-\beta, & r_i^{\mathrm{NC}} < 1 \ \text{or} \ r_i^{\mathrm{DAC}} < 1, \\
\max(0, \hat{A}_i), & \text{otherwise},
\end{cases}
\end{equation}
where $\beta$ is a positive penalty constant. The resulting advantage is used to update
the denoising policy:
\begin{equation}
  \ell_i =
  \sum_{t=1}^{T}
  \gamma^{t-1}
  \log \pi_{\theta}
  \left(
  \tau_{t-1}^{i}
  \mid
  \tau_{t}^{i}, \hat{F}_{t+1:t+f}
  \right),
  \end{equation}

  \begin{equation}
  \mathcal{L}_{\mathrm{GRPO}}
  =
  -\frac{1}{G}
  \sum_{i=1}^{G}
  A_i \ell_i .
  \end{equation}

where $\gamma$ discounts early denoising steps and $\hat{F}_{t+1:t+f}$ denotes the
predicted future occupancy sequence provided by the world model. This reward design
encourages the planner to improve task progress and comfort while explicitly penalizing
unsafe or infeasible trajectories during diffusion-based exploration.

\input{tables/closed}
\section{Experiment}

\subsection{Experimental Setup}

\subsubsection{Dataset}
We conduct experiments on two widely adopted real-world autonomous driving benchmarks: 
nuScenes\cite{caesar2020nuscenes} and NAVSIM\cite{dauner2024navsim}. nuScenes is a multimodal benchmark featuring a 360° sensor suite including cameras, LiDAR, and radar. It contains 1,000 driving scenes with high-quality 3D annotations across 23 object categories, offering greater scale and diversity than previous datasets. NAVSIM offers a data-driven, non-reactive simulation framework that bridges open-loop evaluation and closed-loop performance metrics. It offers curated challenging scenarios (e.g., navtrain and navtest splits), enabling scalable benchmarking of driving policies with consistent closed-loop performance.
\subsubsection{Implementation Details}
Following TransFuser\cite{TransfuserKashyapChitta}, we employ ResNet-34 as the backbone. In open-loop experiments, the diffusion model is trained for 10 epochs while the scene prediction module uses pre-trained weights. We configure the AdamW optimizer with a batch size of 48. The learning rates are configured as $3\times10^{-4}$ for the diffusion and perception modules and $3\times 10^{-5}$ for the scene prediction module. The experiments on NAVSIM were trained for 100 epochs. The learning rates are $6\times10^{-4}$ for the diffusion module and $6\times10^{-5}$ for the scene prediction module. All experiments are implemented on 8 NVIDIA L20 GPUs.

\noindent\textbf{Evaluation metric}
We evaluate planning performance using a comprehensive set of metrics. In the open-loop validation on nuScenes, we calculate the L2 distance between predicted and ground-truth trajectories, which serves as a measure of accuracy. The collision rate quantifies safety by detecting intersections with other agents' bounding boxes, with the average reflecting overall reliability. In the closed-loop validation on the NAVSIM benchmark, we use the Planning-Driven Metric Score (PDMS), which considers No Collision (NC), Driving Ability Check (DAC), Time to Collision (TTC), Comfort (Comf.), and Episode Progress (EP), to assess safety, rule compliance, risk aversion, trajectory smoothness, and navigation efficiency.
\subsection{Comparison Experiments}

\begin{figure*}[t]
    \centering
    \includegraphics[width=1\textwidth]{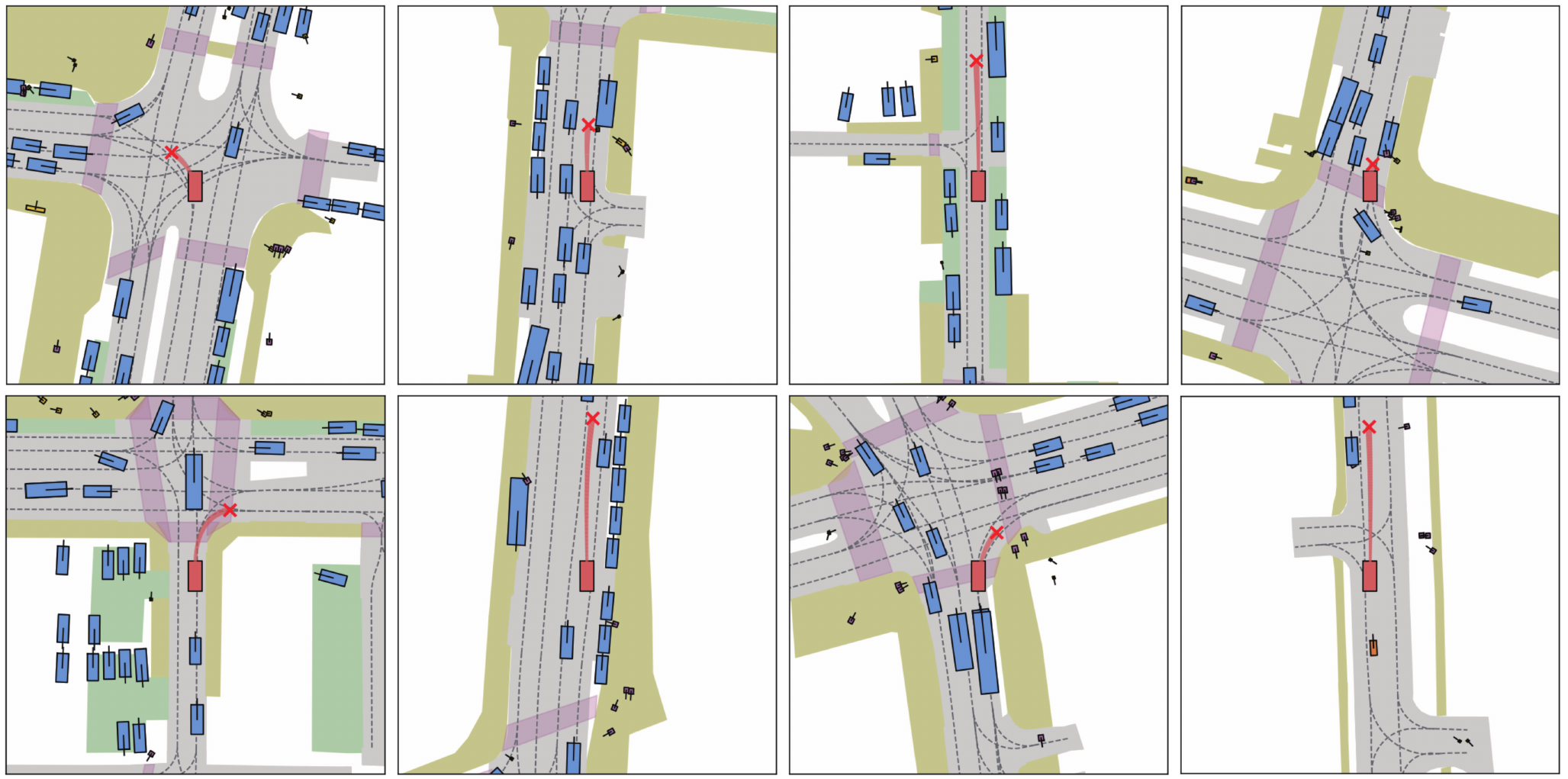}
    \caption{NavSim closed-loop planning visualization. It covers scenarios including going straight, turning left, turning right, overtaking, lane changing, and yielding.}
    \label{fig:full_algorithm}
    \vspace{-15pt}
\end{figure*}
\newcommand{\arraysep}{\renewcommand\arraystretch{1.2}}

\input{tables/abl_nus}
\input{tables/abl_nav}

\input{tables/abl_rl}

\noindent\textbf{Open-Loop Results}
As shown in Table~\ref{tab:nuscenes}, our method achieves superior performance on the nuScenes validation set under open-loop settings. Our approach achieves the lowest average L2 distance of 0.18 m and the lowest collision rate of 0.05 \% across 1 s, 2 s, and 3 s horizons. With respect to L2 distance, our method outperforms all baselines across 1 s, 2 s, and 3 s planning horizons, yielding L2 distances of 0.12 m, 0.18 m, and 0.25 m, respectively. This demonstrates a significant improvement in long-term trajectory accuracy, highlighting the benefit of foresighted reasoning enabled by future scene prediction. Similarly, our method maintains the lowest collision rates at all timesteps, indicating enhanced safety. The results suggest that explicitly modeling temporal dynamics and causal relationships via the learned world model effectively mitigates error accumulation over time, which is a common limitation of reactive planners.

\noindent\textbf{Closed-Loop Results}
As shown in Table~\ref{tab:navsim}, our method achieves the highest overall PDMS score of 90.8, excelling in critical metrics such as a DAC of 97.6 and an EP of 87.3.
The superior DAC score reflects robust decision-making in complex interactions, attributable to the policy distribution refined iteratively by the diffusion planner. Importantly, unlike models relying solely on current observations (e.g., UniAD) or geometric representations (e.g., PARA-Drive), our integration of implicit scene evolution ("Imp") enables more coherent long-horizon planning under occlusion and uncertainty, as evidenced by high EP. The results are consistent with our design objective: leveraging multi-step future prediction to produce trajectories that are causally aligned with anticipated scene dynamics, thereby reducing reliance on heuristic rules.

\subsection{Ablation Study}
\noindent\textbf{Impact of Future Scene Prediction Horizon.}
Integrating future scene information is a core component of our approach. To evaluate its contribution, we systematically vary the number of future frames used to condition the diffusion planner. The results, presented in Table 3 and Table 4, show that planning performance consistently improves as the foresight horizon is extended from 0 s to 3 s.

As shown in Table 3, the baseline model, which relies solely on current and past observations (0 s of future prediction), achieves an average L2 distance of 0.57 m and a collision rate of 0.08 \%. Incorporating 1 s of future prediction yields a noticeable improvement, lowering the average L2 distance to 0.49 m and the collision rate to 0.07 \%. More substantial gains are observed as the horizon is extended to 2 s (0.35 m L2, 0.06 \% collision) and further to 3 s. When conditioned on 3 s of predicted future scenes, the full model reduces the average L2 error by 56 \% and the collision rate by 37.5 \% compared to the baseline, achieving 0.25 m and 0.05 \%, respectively.

These trends are further corroborated by the closed-loop evaluation on the NAVSIM benchmark in Table 4. 
Metrics including DAC and EP show clear improvements as the future horizon increases. The full 3 s model achieves the best overall performance, with a PDMS of 88.4, compared to 87.6 for the baseline.

\noindent\textbf{Impact of Reinforced Diffusion.}
In Table 5, we compare the impact of incorporating the reinforcement module on the results. It can be observed that both DAC and EP have achieved effective improvements, with EP in particular showing a significant enhancement, ultimately resulting in a notable increase in PDMS score.

Overall, the results indicate that extending the temporal context is essential for allowing the planner to anticipate dynamic scene evolution and produce causally consistent long-horizon trajectories with enhanced safety. This indicates that the performance improvement of our method comes from its ability to reason ahead over the next few seconds.


\section{CONCLUSIONS}
We present OWMDrive, a generative end-to-end autonomous driving framework that unifies occupancy-based world modeling and diffusion-based planning. The world model predicts multi-step future 3D occupancy evolution from historical observations and provides these predictions as a conditional prior to the diffusion planner, enabling causality-aware end-to-end trajectory generation under complex interactions and partial observability. Extensive open-loop and closed-loop evaluations show that incorporating future scene rollouts significantly improves planning reliability and safety, especially under challenging conditions such as occlusions and unexpected events. 

However, several limitations remain. The planner still depends on the accuracy of the occupancy world model, and prediction errors may propagate to trajectory generation in highly uncertain or long-tail cases. In addition, the causal dependencies are modeled implicitly through future scene rollouts rather than explicit causal intervention, which may limit interpretability in complex multi-agent interactions. Future work will focus on improving the reliability and uncertainty awareness of the world model, as well as validating the framework in more reactive and safety-critical driving scenarios.









\bibliographystyle{IEEEtran}
\bibliography{References}
\addtolength{\textheight}{-12cm} 
\end{document}

%% file: tables/open.tex
\begin{table*}[t]
\renewcommand{\arraystretch}{1}
\setlength{\tabcolsep}{1.2\linewidth}
\caption{\textbf{Comparisons results with latest methods on the nuScenes\cite{caesar2020nuscenes} dataset with open-loop metrics.} 
}
\centering
      \setlength{\tabcolsep}{8pt}
\fontsize{12}{14}\selectfont
\resizebox{1\textwidth}{!}{
\begin{tabular}{l|cc|ccc>{\columncolor{gray!30}}l|ccc>{\columncolor{gray!30}}l}
\toprule
\multirow{2}{*}{Method} & \multirow{2}{*}{Input}  & \multirow{2}{*}{Scene Representation} &
\multicolumn{4}{c|}{L2 (m) $\downarrow$} & 
\multicolumn{4}{c}{Collision Rate (\%) $\downarrow$} 
\\
& & &  1s & 2s & 3s & Avg. & 1s & 2s & 3s & Avg.\\
\midrule
UniAD~\cite{uniad} & C & Det \& Map \&  Occ \& Motion & 0.44 & 0.67 & 0.96 & 0.69 & 0.04 & 0.08 & 0.23 & 0.12 \\
GenAD~\cite{zheng2024genad} & C & Det \& Map \& Motion & 0.28 & 0.58 & 0.96 & 0.52 & 0.08 & 0.14 & 0.34 & 0.19  \\
BEV-Planner~\cite{ego}& C & None & 0.28 & 0.42 & 0.68 & 0.46 & 0.04 & 0.37 & 1.07 & 0.49 \\
VAD~\cite{vad} & C & Det \& Map \& Motion & 0.41 & 0.70 & 1.05 & 0.72 & 0.07 & 0.17 & 0.41 & 0.22 \\
PARA-Drive~\cite{paradrive} & C & Det \& Map \& Motion & 0.25 & 0.46 & 0.74 & 0.48 & 0.14 & 0.23 & 0.39 & 0.25\\
OccWorld~\cite{zheng2024occworld} & C & Occ & 0.39 & 0.73 & 1.18 & 0.77 & 0.11 & 0.19 & 0.67 & 0.32 \\
SparseDrive~\cite{sun2025sparsedrive} & C & Det \& Map \& Motion & 0.29 & 0.58 & 0.96 & 0.61 & 0.05 & 0.18 & 0.34 & 0.18 \\
DiffusionDrive~\cite{liao2025diffusiondrive} & C  & Det \& Map \& Motion & 0.27 & 0.54 & 0.90 & 0.57 & 0.03 & 0.05 & 0.16 & 0.08 \\
\midrule
\textbf{OWMDrive (ours)} & C & Det \& Map \& Motion &  \textbf{0.12} &  \textbf{0.18} & \textbf{0.25} & \textbf{0.18}  & \textbf{0.01} & \textbf{0.04} & \textbf{0.11} & \textbf{0.05}  \\
\bottomrule
\end{tabular}%
}
\label{tab:nuscenes}
\end{table*}

%% file: tables/closed.tex
\begin{table*}[t]
\setlength{\tabcolsep}{0.01\linewidth}
\caption{\textbf{Comparison on the NAVSIM\cite{dauner2024navsim} navtest split with closed-loop metrics.}
We use \textbf{bold} for the best. }
\centering
\fontsize{12}{14}\selectfont
\resizebox{1\textwidth}{!}{
\begin{tabular}{l|c c|c c c c c|>{\columncolor{gray!30}}c}
\toprule
Method & Input & Scene Representation &
NC$\uparrow$ & DAC$\uparrow$ & TTC$\uparrow$ & Comf.$\uparrow$ & EP$\uparrow$ &
PDMS$\uparrow$ \\
\midrule
Hydra-MDP-$\mathcal{V}_{8192}$~\cite{hydra} & C \& L & Det \& Map \& Motion & 97.9 & 91.7 & 92.9 & \textbf{100} & 77.6 & 83.0 \\

LTF~\cite{TransfuserKashyapChitta}                    & C & Det \& Map \& Occ    & 97.4 & 92.8 & 92.4 & \textbf{100} & 79.0 & 83.8 \\
UniAD~\cite{uniad} & C & Det \& Map \& Occ \& Motion  & 97.8 & 91.9 & 92.9 & \textbf{100} & 78.8 & 83.4 \\
VADv2-$\mathcal{V}_{8192}$~\cite{vadv2} & C & Det \& Map \& Motion & 97.2 & 89.1 & 91.6 & \textbf{100} & 76.0 & 80.9 \\
Transfuser~\cite{TransfuserKashyapChitta}            & C \& L & Det \& Map \& Seg \& Depth & 97.7 & 92.8 & 92.8 & \textbf{100} & 79.2 & 84.0 \\
PARA-Drive~\cite{paradrive}      & C & Map \& Occ \& Motion   & 97.9 & 92.4 & 93.0 & 99.8  & 79.3 & 84.0 \\
DiffusionDrive~\cite{liao2025diffusiondrive}   & C \& L & Det \& Map  & 98.2 & 96.2 & 94.7 & \textbf{100} & 82.2 & 88.1 \\
Hydra-MDP-$\mathcal{V}_{8192}$-W-EP~\cite{hydra} & C \& L & Det \& Map \& Motion & \textbf{98.3} & 96.0 & 94.6 & \textbf{100} & 78.7 & 86.5 \\
DRAMA~\cite{drama}               & C \& L & Det \& Map & 98.0 & 93.1 & \textbf{94.8} & \textbf{100} & 80.1 & 85.5 \\
\midrule
\textbf{OWMDrive (ours)}   & C \& L & Det \& Map \& Motion   & 98.2 & \textbf{97.6} & 94.6 & 99.8 & \textbf{87.3} & \textbf{90.8} \\
\bottomrule
\end{tabular}
}
\label{tab:navsim}
\end{table*}

%% file: tables/abl_nus.tex
\begin{table}[t]
\setlength{\tabcolsep}{0.021\linewidth}
\caption{\textbf{Effect of the foresight scene information forend-to-end autonomous driving. FS} denotes the foresight scene information.
}
\setlength{\tabcolsep}{4.5pt}
\resizebox{0.5\textwidth}{!}{
\begin{tabular}{ccc|cccc|cccc}
\toprule
\multicolumn{3}{c|}{FS} & 
\multicolumn{4}{c|}{L2 (m) $\downarrow$} & 
\multicolumn{4}{c}{Collision Rate (\%) $\downarrow$}  \\
1s & 2s & 3s & 1s & 2s & 3s & \cellcolor{gray!30}Avg. & 1s & 2s & 3s & \cellcolor{gray!30}Avg.  \\
\midrule
$\times$& $\times$ & $\times$ & 0.29 & 0.57 & 0.93 & \cellcolor{gray!30}0.59 & 0.03 & 0.07 & 0.17 & \cellcolor{gray!30}0.09	  \\

$\checkmark$ & $\times$& $\times$ & 0.21 & 0.41 & 0.53  & \cellcolor{gray!30}0.38 & 0.02 & 0.05 & 0.15 & \cellcolor{gray!30}0.07  \\

$\checkmark$ & $\checkmark$& $\times$ & 0.17 & 0.22 & 0.31  & \cellcolor{gray!30}0.23 & \textbf{0.01} & \textbf{0.04} & 0.13 & \cellcolor{gray!30}0.06  \\

$\checkmark$ & $\checkmark$ &$\checkmark$ & \textbf{0.12} & \textbf{0.18} & \textbf{0.25} & \cellcolor{gray!30}\textbf{0.18} & \textbf{0.01} & \textbf{0.04} & \textbf{0.11} & \cellcolor{gray!30}\textbf{0.05}  \\
\bottomrule
\end{tabular}%
}
\label{tab:abl_nuscenes}
\vspace{-5pt}
\end{table}

%% file: tables/abl_nav.tex
\begin{table}[t]
\setlength{\tabcolsep}{0.021\linewidth}
\caption{\textbf{Effect on the NAVSIM closed-loop benchmark without reinforcement learning.}}
\centering
\resizebox{0.5\textwidth}{!}{
\begin{tabular}{ccc|ccccc|>{\columncolor{gray!30}}c}
\toprule
\multicolumn{3}{c|}{FS}& \multicolumn{5}{c|}{Planning Metric} & \\
 1s&2s&3s& NC$\uparrow$ & DAC$\uparrow$ & TTC$\uparrow$ & Comf.$\uparrow$ & EP$\uparrow$ & PDMS$\uparrow$\\
\midrule
$\times$ & $\times$ & $\times$ & 98.2 & 95.8 & \textbf{94.6} & \textbf{100} & 81.9 & 87.6 \\
$\checkmark$ & $\times$ & $\times$ & 98.2 & 96.1 & \textbf{94.6} & \textbf{100} & 82.1 & 88.0 \\
$\checkmark$ & $\checkmark$ & $\times$ &98.2 & 96.5 & 94.5 & 99.9 & 82.5 & 88.3 \\
$\checkmark$ & $\checkmark$ & $\checkmark$ &98.2 & \textbf{96.6} & 94.5 & \textbf{100} & \textbf{82.6}& \textbf{88.4} \\
\bottomrule
\end{tabular}
}
\label{tab:abl_navsim}
\vspace{-8pt}
\end{table}

%% file: tables/abl_rl.tex
\begin{table}[t]
\setlength{\tabcolsep}{0.021\linewidth}
\caption{\textbf{Effect of reinforced diffusion.}}
\centering
      \setlength{\tabcolsep}{4pt}
\resizebox{0.5\textwidth}{!}{
\begin{tabular}{c|ccccc|>{\columncolor{gray!30}}c}
\toprule
\multirow{2}{*}{Reinforcement}&\multicolumn{5}{c|}{Planning Metric} & \\
 &NC$\uparrow$ & DAC$\uparrow$ & TTC$\uparrow$ & Comf.$\uparrow$ & EP$\uparrow$ & PDMS$\uparrow$\\
\midrule
$\times$&98.2 & 96.6 & 94.5 & \textbf{100} & 82.6& 88.4 \\
$\checkmark$&98.2 & \textbf{97.6} & \textbf{94.6} & 99.8 & \textbf{87.3} & \textbf{90.8}\\
\bottomrule
\end{tabular}
}
\label{tab:abl_navsim}
\vspace{-8pt}
\end{table}